\documentclass[journal,twocolumn]{IEEEtran}
\usepackage{amsfonts}
\usepackage{times}
\usepackage{graphicx}
\usepackage{latexsym}
\usepackage{dsfont}
\usepackage{amssymb}
\usepackage{amsmath}
\usepackage{cite}
\usepackage{verbatim}

\newcommand{\figref}[1]{{Fig.}~\ref{#1}}

\def\bb0{{\mathbb{0}}}

\def\ba{{\mathbf{a}}}
\def\bb{{\mathbf{b}}}

\def\b0{{\mathbf{0}}}

\def\sf0{{\mathsf{0}}}

\usepackage[table]{xcolor}
\usepackage{epstopdf}
\usepackage{makecell}
\usepackage{array,colortbl,multirow,multicol,booktabs,ctable}
\usepackage{enumerate}
\usepackage{algorithmicx}
\usepackage{algorithm}
\usepackage{amsmath}
\usepackage[noend]{algpseudocode}
\usepackage{float}
\usepackage{soul}
\usepackage{hyperref}
\usepackage{makeidx}
\usepackage{bbm}
\usepackage{graphicx}
\usepackage[font=small]{caption}
\usepackage{array}
\usepackage{epsfig}
\usepackage{comment}
\usepackage{textcomp}
\usepackage{subcaption}
\usepackage{nopageno}
\usepackage{cite}

\begin{document}
\bstctlcite{IEEEexample:BSTcontrol} 
\title{Deep Learning for Moving Blockage Prediction using Real Millimeter Wave Measurements}

\author{\IEEEauthorblockN{Shunyao Wu, Muhammad Alrabeiah, Andrew Hredzak, Chaitali Chakrabarti, and Ahmed Alkhateeb}\\
\IEEEauthorblockA{\textit{School of Electrical, Computer and Energy Engineering}\\
\textit{Arizona State University Tempe, AZ 85287}\\
Email: \{vincentw, malrabei, ahredzak, chaitali, alkhateeb\}@asu.edu}
}

\maketitle

\begin{abstract}
Millimeter wave (mmWave) communication is a key component of 5G and beyond.  Harvesting the gains of the large bandwidth and low latency at mmWave systems, however, is challenged by the sensitivity of mmWave signals to blockages; a sudden blockage in the line of sight (LOS) link leads to abrupt disconnection, which affects the reliability of the network.  In addition, searching for an alternative base station to re-establish the link could result in needless latency overhead. In this paper, we address these challenges collectively by utilizing machine learning to anticipate dynamic blockages proactively. The proposed approach sees a machine learning algorithm learning to predict future blockages by observing what we refer to as the \textit{pre-blockage signature}. To evaluate our proposed approach, we build a mmWave communication setup with a moving blockage and collect a dataset of received power sequences. Simulation results on a real dataset show that blockage occurrence could be predicted with more than 85\% accuracy and the exact time instance of blockage occurrence can be obtained with low error. This highlights the potential of the proposed solution for dynamic blockage prediction and proactive hand-off,  which enhances the reliability and latency of future wireless networks. 
\end{abstract}

\thispagestyle{empty}

\section{Introduction} 
\label{sec:Intro}
Communication at the mmWave frequency range offers high bandwidth and higher data rate demands required by 5G and beyond cellular systems \cite{heath2016overview}. However, mmWave systems struggle in the presence of objects that block the LOS connection between a base station and its users \cite{alkhateeb2018machine}. This is majorly rooted into the poor penetration and reflection properties of mmWave signals \cite{heath2016overview,rappaport2017overview}. This struggle with blockages or Non-LOS (NLOS) connections ultimately affects the reliability of the system. In addition to that, it introduces a latency burden resulting from the need for user hand-off \cite{alkhateeb2018machine}. A key approach to address that struggle lies in equipping the mmWave system with the capability to predict possible blockage proactively. Such successful prediction allows a base station to take mitigation measures, e.g., user hand-off, before the link is blocked, thereby resolving the reliability and latency problems.

Many recently published studies have used machine learning to address problems arising from link blockages in MIMO and mmWave communications \cite{choi2017deep,huang2020machine,alkhateeb2018machine,alrabeiah2020deep,charan2020vision}. 
The work in \cite{choi2017deep, huang2020machine, alrabeiah2020deep} collectively demonstrates the ability of a machine learning algorithm (whether deep \cite{huang2020machine} or shallow \cite{choi2017deep}) to identify or differentiate LOS and NLOS links. This identification task could be an interesting ability to a system operating in the sub-6 GHz range.
However, the requirements are more stringent for a mmWave system and demand a proactive approach. A step towards doing so is presented in \cite{alkhateeb2018machine}, where a proactive solution is proposed to predict stationary blockages. This solution depends on beam sequences alone, and as such, it cannot handle dynamic blockages. Another direction addressing blockage prediction relies on Vision-Aided Wireless Communications \cite{CamBeamPred,charan2020vision}. In \cite{CamBeamPred}, a camera feed along with sub-6 GHz channels are used to identify currently blocked mmWave links. This work is the seed to that presented in \cite{charan2020vision}, where proactive blockage prediction using images and mmWave beams is attempted using bimodal deep learning algorithm. Despite the appeal, that work require extra sensory data (images), and it does not take full advantage of the wireless data.

In this paper, we propose a machine learning algorithm that addresses proactive dynamic-blockage prediction. The algorithm uses sequences of received powers to predict whether a blockage is incoming or not. The basic idea behind our algorithm is the ability to recognize what we have dubbed \textit{pre-blockage signature}. We argue that signature could serve as a important clue for incoming blockages. Our contribution is summarized as follows:
\begin{enumerate}
\item We propose a recurrent neural network (RNN) architecture based on Gated-recurrent units (GRUs) to predict incoming blockages. The architecture is designed to learn one of two tasks. The first is to predict whether a blockage is incoming or not, and the other is to pinpoint the time instance at which the blockage occurs.
\item We develop a mmWave communication setup with a moving blockage. We use that setup to build a dataset of received power sequences and their corresponding future link statuses. The dataset is utilized to the proposed algorithm and its ability to predict incoming blockages.
\end{enumerate}

The rest of this paper is organized as follows. Section \ref{sec:sys_ch} presents the system and channel models adopted to study the dynamic-blockage prediction. Section \ref{sec:prob_formul} presents a formulation of the prediction problem. The proposed machine learning model is presented in Section \ref{sec:deep_learning}. The data collection scenario and setup is introduced in Section \ref{sec:experimental}. Proposed algorithm evaluation and simulation results are presented in Section \ref{sec:results}, and, finally, the paper is concluded in Section \ref{sec:con}.

\section{System and Channel Models}
\label{sec:sys_ch}
The following two subsections will introduce the system and channel models adopted in this paper.
\begin{figure}[t]
	\centering
	\includegraphics[width=\linewidth]{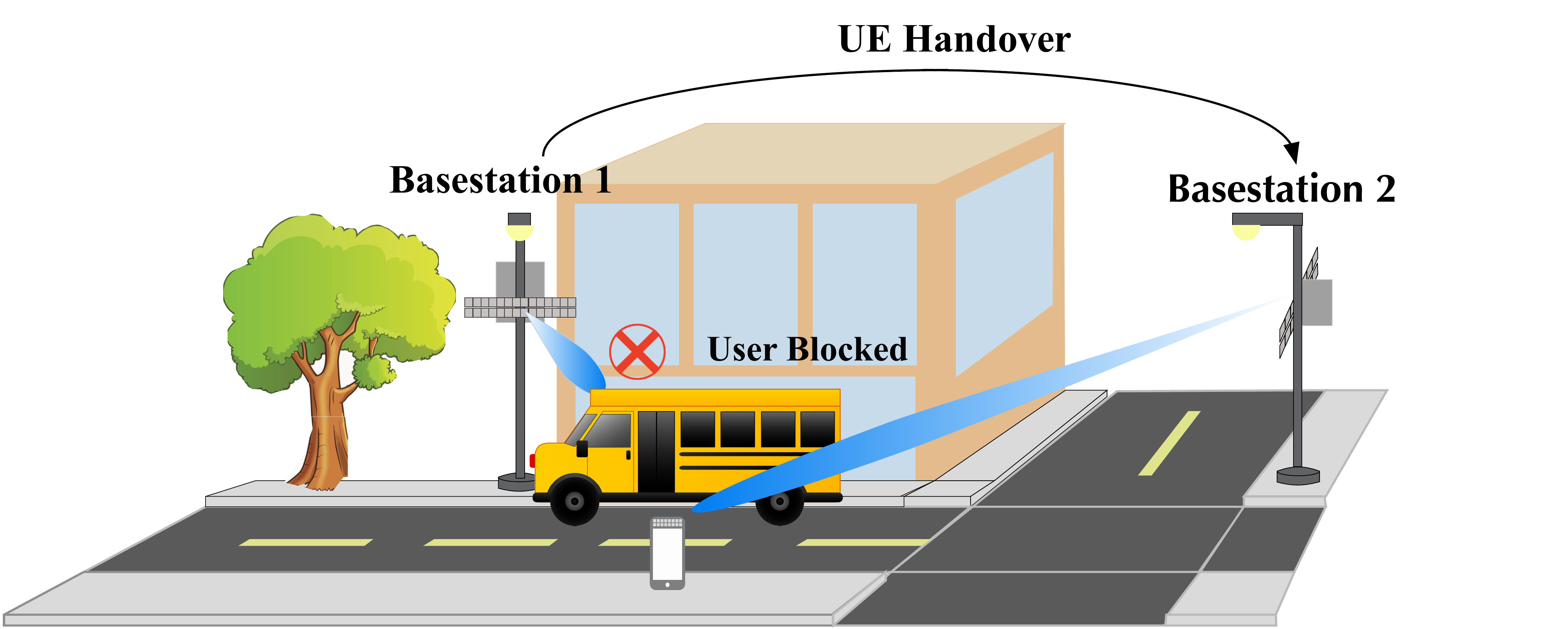}
	\caption{An illustration of the considered system model.}
	\label{fig:system diagram}
\end{figure}

\textbf{System model:}
The communication system considered in this work is described in Fig.\ref{fig:system diagram}. It assumes a base station serving a static user who is in the vicinity of a possible moving blockage.   The base station employs an $M$-element Uniform Linear Array (ULA) antenna operating at 60GHz carrier frequency with Orthogonal Frequency Division Multiplexing (OFDM). It also assumes a fully analog architecture with a predefined beam-steering codebook $\mathcal F = \{\mathbf f_w\}_{w=1}^{W}$, where $\mathbf f_w\in \mathbb C^{M\times 1}$ is given by: 
\begin{equation}\label{eq:array_steering}
	\mathbf f_w = \frac{1}{\sqrt{M}}\left[ 1, e^{j\frac{2\pi}{\lambda}d\sin(\phi_w)},\dots, e^{j(M-1)\frac{2\pi}{\lambda}d\cos(\phi_w)} \right]^T,
\end{equation}
where $d$ is the spacing between the ULA elements, $\lambda$ is the wavelength, and $\phi_w \in\{\frac{2\pi w}{W}\}_{w=0}^{W-1}$ is a uniformly quantized azimuth angle with a step of $1/W$. At any time instance $t$, the downlink received signal is expressed as follows:

\begin{equation}\label{sig_model}
    r_k[t] = \mathbf h_k[t]^T \mathbf f_w s_k[t] + n_k
\end{equation}
where $\mathbf h_k[t]\in \mathbb C^{M\times 1}$ is the downlink channel at the $k$th subcarrier, $s_k[t]$ is the symbol transmitted on the $k$th subcarrier, and, finally, $n_k$ is a complex Gaussian noise sample, $\sim \mathcal {CN}(0,\sigma^2)$ at the $k$th subcarrier. 

\textbf{Channel model:}
This work adopts the geometric (physical) channel model, which captures the physical characteristics  of signal propagation including the dependence on the environment geometry, materials, frequency band, etc., \cite{heath2016overview}. With this model, the channel can be expressed as: 
\begin{equation}
\mathbf h_k = \sum_{d=0}^{D-1} \sum_{\ell=1}^L \alpha_\ell e^{- \j \frac{2 \pi k}{K} d} p\left(dT_\mathrm{S} - \tau_\ell\right) \ba\left(\theta_\ell, \phi_\ell\right),
\end{equation} 
where $L$ is number of channel paths, $\alpha_\ell$ is the path gain (including the path-loss), $\tau_\ell$ is the delay, $ \theta_\ell$ is the azimuth angle of arrival, and $\phi_\ell$ is the elevation angle, of the $\ell$th channel path. $T_\mathrm{S}$ represents the sampling time while $D$ denotes the cyclic prefix length (assuming that the maximum delay is less than $D T_\mathrm{S}$).

\section{Problem Formulation} 
\label{sec:prob_formul}
Proactively identifying Line of Sight (LOS) link status has significant advantages both at the physical and network layer levels. In this paper, we focus on two specific problems: (i) How to use the received mmWave signal power information to predict whether there is a blockage in the near future or not, and (ii) in case there is a blockage, how to use the received mmWave signal power information to predict when that blockage will occur. 

\textbf{{Problem 1:}} To formulate the presence of a blockage in the near future, let $t\in \mathbb Z$ be the index of the discrete time instance, $x[t]$ be the link status at the $t$th time instance, and let $S_{ob} = \{ \lvert r[t+n] \rvert ^2 \}_{n=-T_{o}+1}^{0}$ be the sequence of received signal power for the observation interval of $T_{o}$ instances. Note that for simplicity of expression, the subcarrier index $k$ is dropped from $r[t+n]$. Given a signal power based observation sequence, we want to predict the occurrence of blockage within a future time interval extending over $T_P$ instances. We use $b_{T_P}$ to indicate whether there is a blockage occurrence within that interval or not. More formally, $b_{T_p}$ is defined as follows:
\begin{equation} \label{equ:p1_label}
    b_{T_P} = 
    \begin{cases}
      0, &  x[t+n^{\prime}] = 0 \quad \forall n^{\prime} \in \{ 1,\dots,T_P \} \\
      1, & \text{otherwise}
    \end{cases}       
\end{equation}
where $1$ indicates the occurrence of a blockage and $0$ is the absence of blockage. The goal of this problem is to predict $b_{T_p}$ with high accuracy, i.e., high success probability $\mathbb P( \hat {b}_{T_p}= b_{T_p} | S_{ob})$ where $\hat {b}_{T_p}$ is the predicted link status. To that end, a prediction function $f_{\Theta}(S_{ob})$ parameterized by a set of parameters $\Theta$ could to be learned using a machine learning algorithm such that it 
maximizes $\mathbb P( \hat {b}_{T_p} = b_{T_p}| S_{ob})$.

\begin{figure*}[t]
	\centering
	\includegraphics[width=\linewidth]{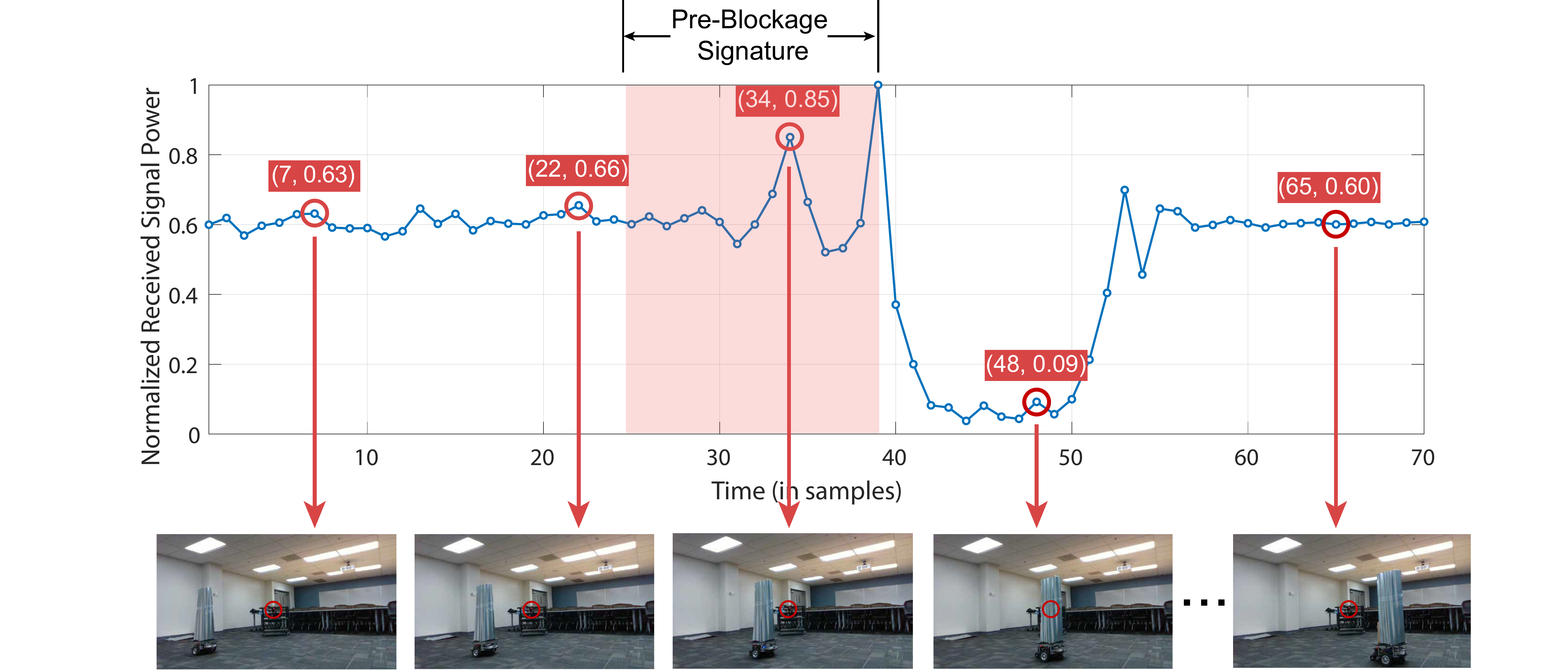}
	\caption{Example of received signal power vs time. It also shows photos of the location of the transmitter and blockage (from the receiver perspective).}
	\label{fig:ori_ch_pow}
\end{figure*}

\textbf{{Problem 2:}} Given signal power based observation sequence $S_{ob}$ and the knowledge that there is a blockage in the future $T_p$ instances, the goal is to predict $n^{\prime}$ at which $ x[t+n^{\prime}] = 1$. This represents the exact instance at which the blockage occurs within a window of $T_p$ instances. Similar to {\textbf{Problem 1}}, the future instance is predicted by a parameterized function $g_{\Theta}(\mathcal S_{ob})$ that could be learned using a ML algorithm. The aim of the ML algorithm is to maximize the prediction accuracy $\mathbb P(\hat{n}^{\prime} = n^{\prime}|\mathcal S_{ob}, b_{T_p}=1)$.

\section{Moving Blockage Prediction using Recurrent Neural Networks} 
\label{sec:deep_learning}
Deep neural networks have recently established themselves as a powerful learning algorithm for many applications \cite{resnet}\cite{DenNets}\cite{DLBook}, and as such, they will be the center of the proposed solution for the future-blockage prediction problem.

\subsection{Key Idea} 

Moving objects in a wireless communication environment contribute to changes in the wireless channels, resulting in obvious fluctuations in received signal power. This fluctuation pattern in received signal power is referred to as the \textit{Pre-Blockage Signature}~\cite{DL_coord_beam}. Such pattern is illustrated in Fig.~\ref{fig:ori_ch_pow}. It shows a captured sequence of received power versus time instances, and the corresponding photos show how far the blockages (the metal object in the photos) is from the transmitter (the object circled with red in the photos). The received power starts with smooth fluctuations (between the 1-st and 30-th instances in \figref{fig:ori_ch_pow}), for the blockage is far from both the receiver and the transmitter. However, as that blockage advances, its effect becomes clearer in the received sequence, which could be seen in the red-shaded region of \figref{fig:ori_ch_pow}. The sharper fluctuations is what we refer to as the pre-blockage signature, and it could be utilized to identify incoming blockages.

\begin{figure}[t]
	\centering
	\includegraphics[width=\linewidth]{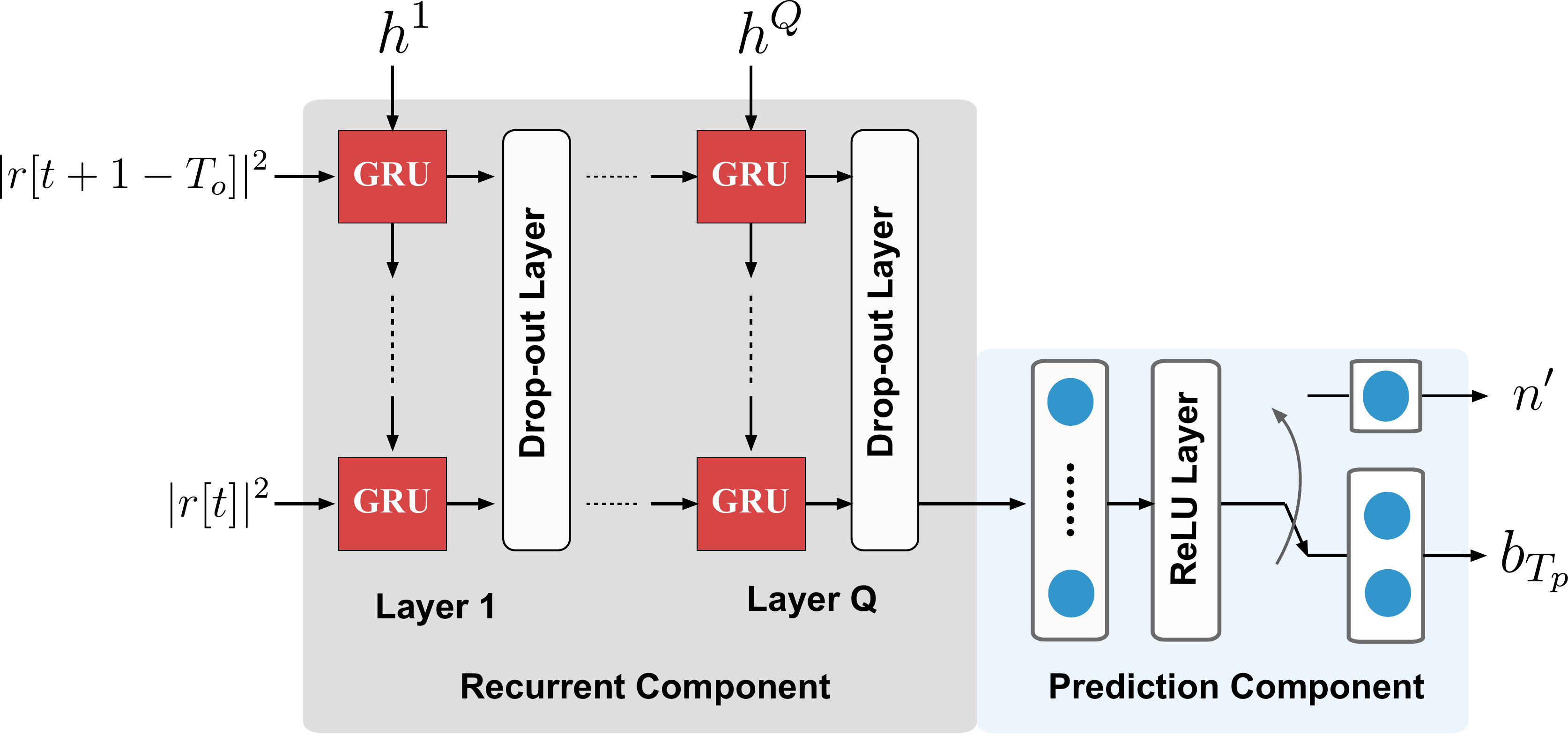}
	\caption{The overall RNN architecture composed of a recurrent and a prediction component}
	\label{fig:NN_arch}
\end{figure}

\subsection{Deep Learning Model}
\label{deep_learning_model}

\textbf{Neural Network Architecture:} Learning the pre-blockage signature from a sequence of observed received signal power requires a neural network that could process input data samples over time such as recurrent neural networks. We design a Gated Recurrent Unit (GRU) network of $Q$-layers that takes in a sequence of observed received signal powers (i.e., $S_{ob}$) and learns to predict the link status $b_{T_p}$. Fig.\ref{fig:NN_arch} depicts the schematic of such a network. Each layer in the network comprises a sequence of GRUs equal to $T_o$. The output of the last GRU of the last layer is fed to a Fully Connected (FC) layer followed by either a classifier for \textbf{{problem 1}} or a regressor for \textbf{{problem 2}}. The classifier outputs a probability vector ($\hat{\mathbf p}$) of whether the link status is blocked or not in $T_P$ future time instances. For \textbf{{problem 2}}, the regressor outputs the predicted time instance $\hat{n}^{\prime}$ indicating the time instance at when the blockage will occur.

\textbf{Pre-Processing:} Before we input the data into our network, we need to pre-process it to make it suitable for our model to learn, see \cite{EffBackProp} for more information. We choose to standardize the inputs by subtracting the mean $\mu$ of the dataset and dividing by its standard deviation $\sigma$. Let $\mathbf A\in\mathbb R^{U\times N}$ be the dataset matrix, where each row represents a data point with a total number of data points of $U$. Data standardization is done by computing:
\begin{equation}
\mathbf {\hat{A}}_{u,n} = \frac{\mathbf {A}_{u, n} - \mu}{\sigma},
\end{equation}
where:
\begin{equation}
\mu = \frac{1}{N \times U} \sum_{u=1}^{U}\sum_{n=1}^{N}\mathbf{A}_{u, n}
\end{equation} 

\begin{equation}
\sigma =\sqrt{ \frac{1}{N\times U}\sum_{u=1}^{U}\sum_{n=1}^{N}(\mathbf{A}_{u, n} - \mu})^2
\end{equation}
\begin{equation}
    \forall u \in \{1,\dots,U\} \text{and } n \in \{1,\dots,N\},    
\end{equation}

\textbf{Training loss:} For \textbf{{problem 1}} the future link-status prediction is posed as a binary classification problem, in which the classifier attempts to determine whether the link is blocked or not within the future time interval. As such, the network training is performed with a cross entropy loss function \cite{DLBook} computed over the outputs of the network:
\begin{equation}
l_{\text{CH}} =  \sum_{c = 1}^{2} p_{c}\log{\hat{p}_{c}},
\end{equation}
where $\mathbf p = [p_1, p_2]^T$ is the one-hot vector where a one hot vector is a representation of categorical variables as binary vector, the category with highest probability is encoded as 1 others are encoded as 0's. It takes one of two values: $[1,0]^T$ for the case when $b_{T_p} = 0$ and $[0,1]^T$ for the case when $b_{T_p} = 1$, and $l_{\text{CH}}$ is the training loss computed for one data point.

For \textbf{{problem 2}}, we pose the problem of predicting the blockage instance as a regression problem. Our model tries to determine the exact time instance at which the blockage occurs. We use Mean Square Error (MSE) loss as training function. In formal terms, we aim to minimize the difference between the predicted instance and groundtruth instance:
\begin{equation}
l_{\text{MSE}} =  (n^{\prime (u)}- \hat{n}^{\prime (u)})^2
\end{equation}
where $n^{\prime (u)}$ and $\hat{n}^{\prime (u)}$ are ground truth time instance and predicted time instance, respectively. 

\section{Experimental Setup and Evaluation Dataset}
\label{sec:experimental}

\subsection{Communication Scenario \& Testbed}
\label{subsec:scenario}

 \begin{figure}[t]
	\centering
	\includegraphics[width=\linewidth]{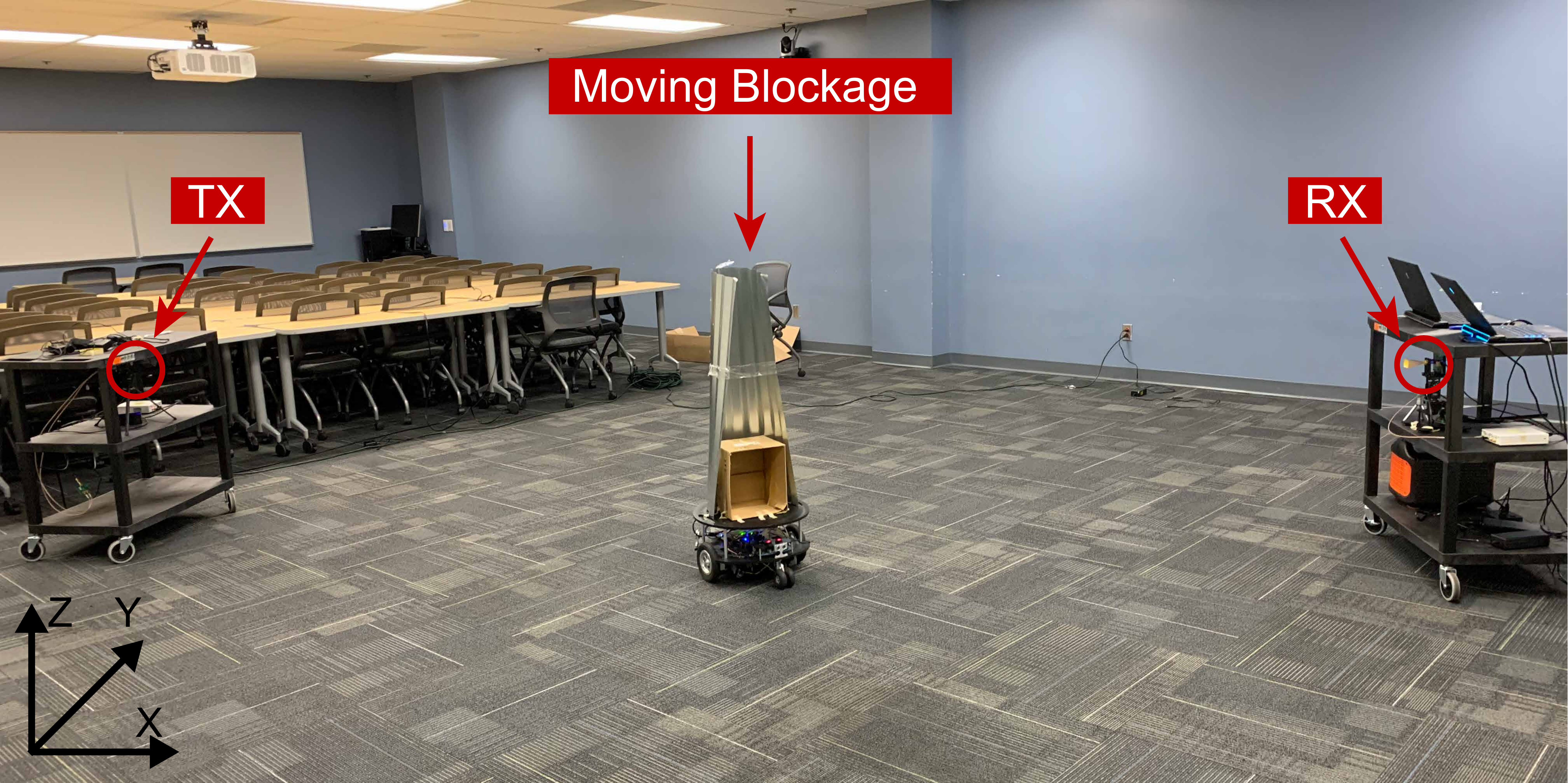}
	\caption{Experimental Scenario. The blockage moves along a trajectory between the transmitter (TX) and receiver (RX).}
	\label{fig:exp_setup}
\end{figure}

\begin{table}[tb]
	\centering
	\small
	\caption{Parameters for mmWave Communication Systems}
	\begin{tabular}{cc}
		\hline\hline
	\textbf{Name} & \textbf{Value} \\ 
	\hline
		Carrier Frequency   & 60GHz  \\ 
		Signal Bandwidth   & 20MHz  \\ 
		number of subcarriers  & 64  \\ 
	    Horn Antenna Gain   & 20dBi \\ 
	    Transmit Power & 30dBm \\ 
	    \hline\hline
	\end{tabular}
	\label{tbl:comm_sys_para}
\end{table}
We build a mmWave communication system comprising of a transmitter with an omnidirectional antenna communicating with a receiver with a 10-degree beamwidth horn antenna. The operating parameters for our mmWave communication system are shown in Table~\ref{tbl:comm_sys_para}. To simulate a moving blockage, we use a metal cylinder with a height of 1 m, which can completely block the LOS link between the transmitter and receiver. Then, we mount that cylinder onto a programmed robot that moves along a pre-defined trajectory to simulate the moving blockage. \figref{fig:exp_setup} depicts our system.

\begin{figure}[t]
	\centering
	\begin{subfigure}{0.49\textwidth}
	\centering
		\includegraphics[width=\linewidth]{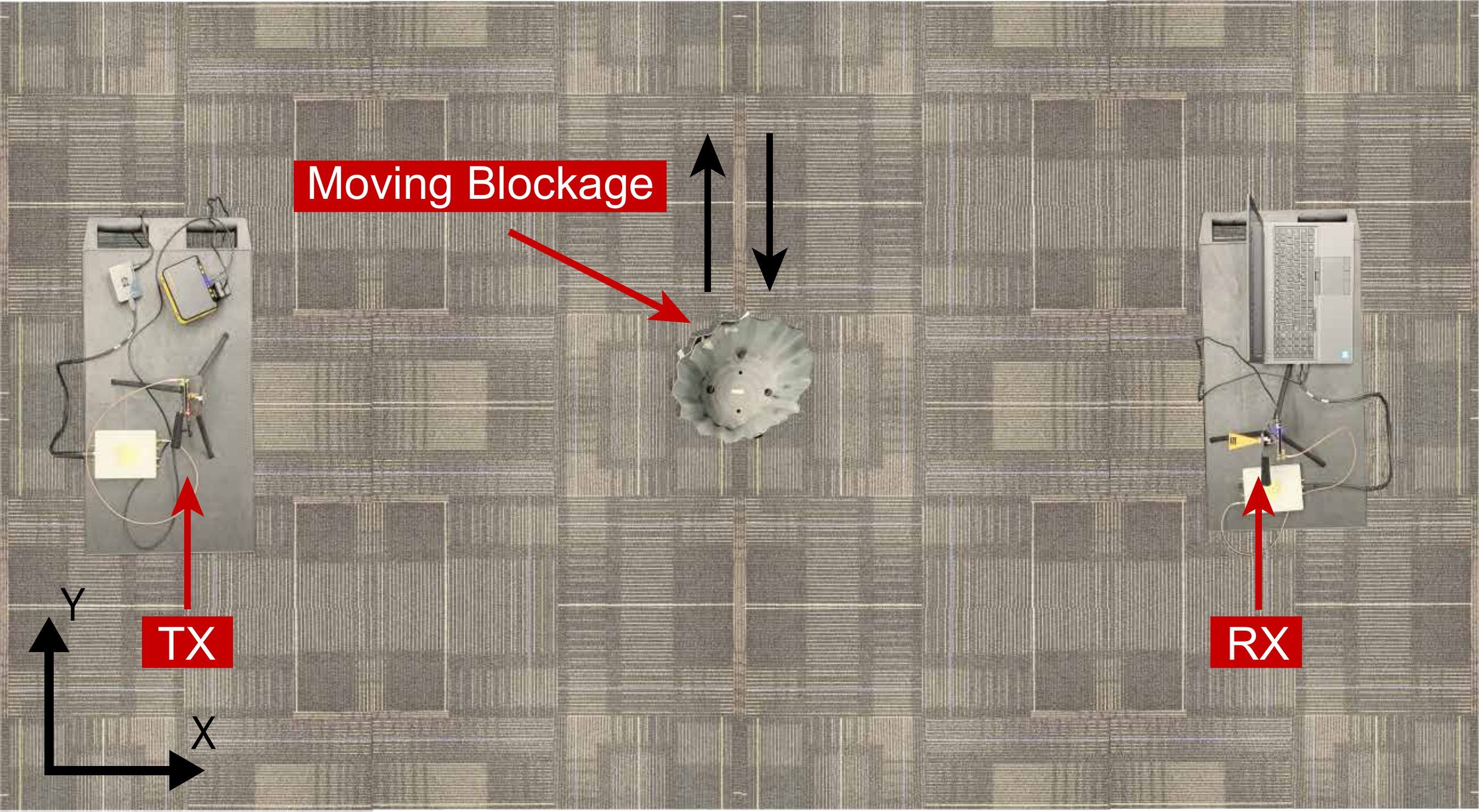}
		\caption{ }
		\label{fig:top_view}
	\end{subfigure}
	\begin{subfigure}{0.49\textwidth}
	\centering
		\includegraphics[width=\linewidth]{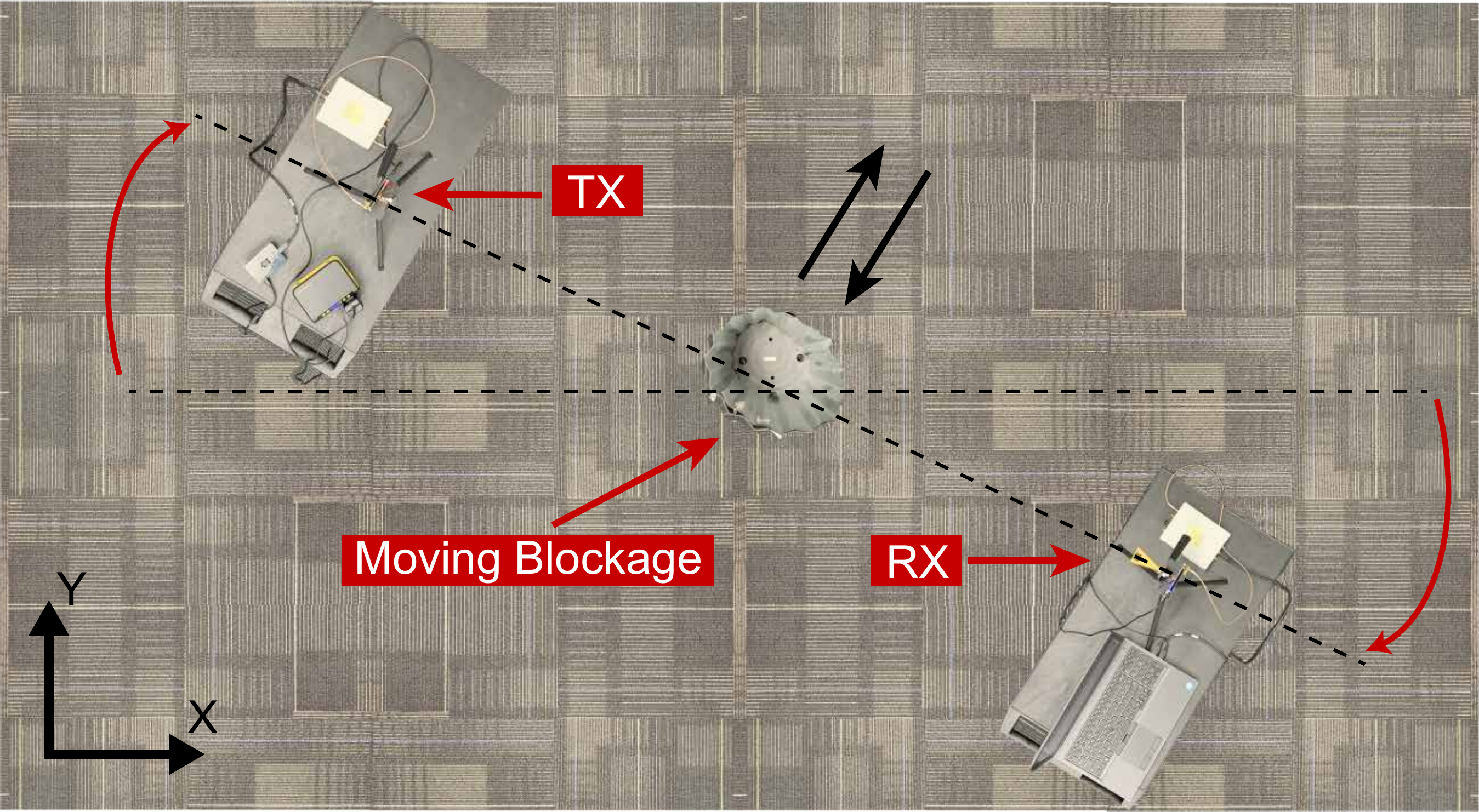}
		\caption{ }
		\label{fig:top_view_rot}
	\end{subfigure}
	\caption{Top-view of the experimental scenario. (a) shows the trajectory of the moving blockage and the relative positions of transmitter (TX), receiver (RX) and moving blockage. (b) shows the rotated TX-RX setup.}
\end{figure}
We consider an indoor scenario where the transmitter and receiver are placed 8 m apart from each other and the robot moves between them in different trajectories. To illustrate this further, let's consider the coordinate system in \figref{fig:exp_setup}, where the x-axis extends along the LOS between the transmitter and receiver and y-axis extends perpendicular to the LOS between the transmitter and receiver. z-axis is perpendicular to the ground. The robot moves back and forth on the y-axis to create multiple back and forward trajectories with a spacing of 1 m. Then, the whole testbed is rotated with a small angle on the x-y plane such that the background is slightly changed, and the robot is programmed to do another back and forth cycle; see \figref{fig:top_view} and \figref{fig:top_view_rot} for an example of the robot motion in the rotated testbed. By continuously moving and rotating, we collect a dataset of power sequences that encodes different motion and propagation patterns.

\subsection{Dataset Generation} 
\label{subsec:dataset}

\begin{figure}[t]
	\centering
	\includegraphics[width=1\linewidth]{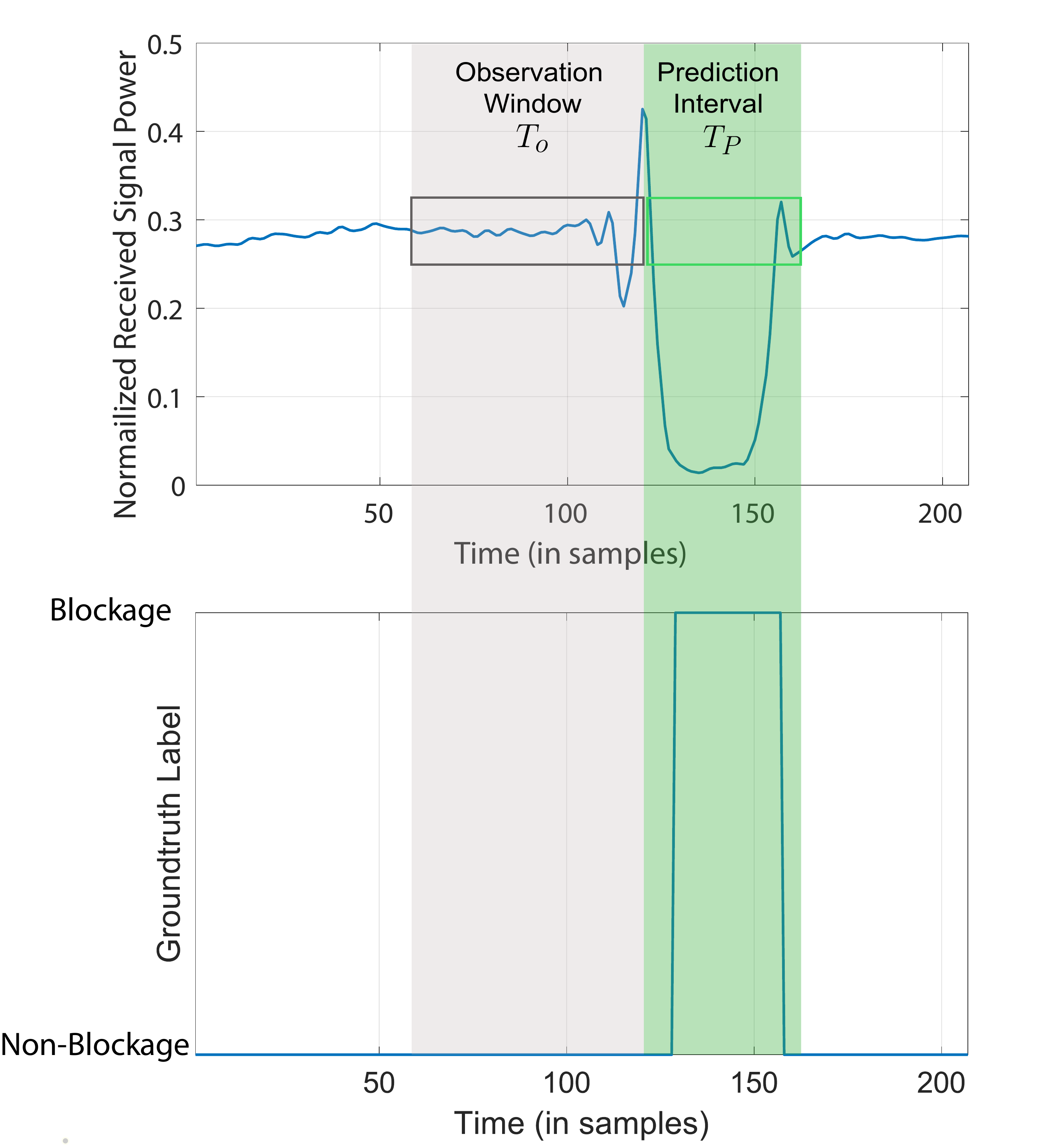}
	\caption{Sequence generation using sliding window. The top image shows the received power for a raw sequence. The bottom image shows the correspond link status.}
	\label{fig:slicing}
\end{figure} 
Every robot trajectory provides us with a single received-power sequence, which is manually annotated to create the link status labels. We use 1 to indicate the time instances at which the LOS link is blocked and 0 otherwise. We call the pair of received power and link status sequence a raw sequence pair. Furthermore, since the number of raw sequence pairs we collected from the experiment is limited, data augmentation is used to increase the dataset size. Originally, we conducted 158 experiments and generated 158 sequence pairs, i.e. $\mathcal S_{1}=\{(S_{d1},x_{d1})^{(u)} \}_{u=1}^{U_d}, U_d = 158$, $d_1$ means the original input. We generate additional pairs by dropping samples at rates 2, 3, and 4, which results in $\mathcal S_{2}=\{(S_{d2},x_{d2})^{(u)} \}_{u=1}^{U_d}$, $\mathcal S_{3}=\{(S_{d3},x_{d3})^{(u)} \}_{u=1}^{U_d}$, $\mathcal S_{4}=\{(S_{d4},x_{d4})^{(u)} \}_{u=1}^{U_d}$, respectively. In reality, this procedure means the blockage moves along the same trajectory at 2,3 or 4 times the original speed. Next we concatenate these sequences together as $\mathcal S=\mathcal S_{1} \bigcup \mathcal S_{2} \bigcup \mathcal S_{3} \bigcup \mathcal S_{4} = \{ (S,x)^{(u)}\}_{u=1}^{U} $ where $U = 4U_d$.

The method to generate the data points for \textbf{Problem 1} and \textbf{Problem 2} are as follows.

\textbf{Problem 1:} A data point consists of an input received signal power sequence $S_{ob}$ and the input label $b_{T_p}$. To generate the received signal power sequence, we use a sliding window method, shown in Fig.~\ref{fig:slicing}. For example, assume that current time is $t$, we generate $S_{ob}$ by extracting the received power sequence from time instance $t-T_o + 1$ to $t$ shown as red box in the top subplot in Fig.~\ref{fig:slicing}. For input label generation, we first extract the label sequence from time instance $t+1$ to $t+T_P$ shown as green box in bottom subplot in Fig.~\ref{fig:slicing}, this gives us sequence $\{x[t+n]\}_{n=t}^{t+T_P}$. Then we generate input labels $b_{t_P}$ using eqn.~\ref{equ:p1_label}. Finally, we pair the received signal power sequences with input labels, expressed as $S_{P1} = \{  (S_{ob},b_{T_P})^{(u)} \}_{u=1}^{U_{P1}}$, where $U_{P1}$ is the total samples we input to our model. In this problem,  $S_{P1}$ are mixed with two categories, non-transition sequence pairs where $b_{T_P} = 0$ and transition sequence pairs where $b_{T_P} = 1$. To eliminate the dataset bias, we keep the ratio of transition and non-transition sequence pairs to 1:1.

\textbf{Problem 2:} A data point is represented by input received signal power sequence $S_{ob}$ with the ground-truth time instance $n^{\prime}$ instead of the label. So $S_{P2} = \{  (S_{ob},n^{\prime })^{(u)} \}_{u=1}^{U_{P2}}$, where $U_{P2}$ is the total number of samples that are input to our model. In this problem, we only select transition-sequence pairs to be our input dataset.

In our experiments, we pick input sequence length ($T_{o}$) as 10 and prediction interval ($T_{p}$) from 1 to 40. For each $T_{p}$, each raw sequence pair generates one transition-sequence pair and multiple non-transition sequence pairs. Totally, we get 632 transition-sequence pairs for each $T_p$.

\section{Experimental Results} 
\label{sec:results}

\subsection{Evaluation Metrics}
\label{subsec:eva_metric}

Since \textbf{{problem 1}} is a classification problem, we use Top-1 accuracy as our evaluation metric. It is defined as:
\begin{equation}
     \text{Acc}_{\text{top-1}} = \frac{1}{U_{v1}}\sum_{u = 1}^{U_{v1}} \mathbbm{1}  (b_{T_p}^{(u)} = \hat {b}_{T_p}^{(u)}),
\end{equation}
where $\mathbbm{1}$ is the indicator function, $U_{v1}$ is total samples of the validation set in problem 1, $b_{T_p}^{(u)}$ and $\hat{b}_{T_p}^{(u)}$ are, respectively, the target and predicted link status for a future interval of $T_P$ instances.

\textbf{{Problem 2}} is posed as a regression problem, and so we use Mean Absolute Error (MAE) and its standard deviation to evaluate the quality of our model predictions. The MAE is the mean absolute error between ground-truth value and predicted value. For each prediction interval $T_P$, we calculate MAE across the samples and standard deviation of these absolute errors. 
\begin{equation}
     e^{(u)}_{T_P} = \lvert n^{\prime (u)} - \hat{n}^{\prime (u)} \rvert, \quad  \forall u \in \{1,\dots,U_{v2}\} ,    
\end{equation}
\vspace{-0.4cm}
\begin{equation}
     \bar {e}_{T_P} = \frac{1}{U_{v2}}\sum_{u = 1}^{U_{v2}} \lvert n^{\prime (u)} - \hat{n}^{\prime (u)}  \rvert,
\end{equation}

\begin{equation}
     \text{std}_{T_P} = {\sqrt {{\frac {1}{U_{v2}}}\sum _{i=1}^{U_{v2}}\left(e^{(u)}_{T_P}-{ \bar {e}_{T_P}}\right)^{2}}},
\end{equation}
where, $e^{(u)}_{T_P}$ is the absolute error for $u$th sample, $\bar {e}_{T_P}$ is the MAE, $ \text{std}_{T_P}$ is the standard deviation of absolute error, $U_{v2}$ is the total numbers of samples in validation set, $n^{\prime (u)}$ and $\hat{n}^{\prime (u)}$ are target and predicted time instances between current time and the time of blockage occurrence, assuming prediction interval is $T_P$.

\subsection{Network Training}
\label{subsec:net_train}
We build the deep learning model described in Section~\ref{sec:deep_learning} using Pytorch. We input 10 successive time-instances of received signal power and the corresponding label for training. Our model consists of 1 RNN layer with 20 hidden states with a dropout rate = 0.2. The number of epochs is 1000. These parameters are chosen based on empirical experiments. The detailed parameters of our RNN for \textbf{{problem 1}} and \textbf{{problem 2}} are shown in Table~\ref{tbl:DP_para}.

\begin{table}[tb]
\centering
\caption{Parameters for Deep Learning Model}
\begin{tabular}{ccc}
\cline{1-3}
\hline\hline
 & \multicolumn{2}{c}{ \textbf{Value}}  \\ 
\multirow{-2}{*}{\textbf{Name}} &  \textbf{Problem 1}    &  \textbf{Problem 2}     \\
\hline
Input sequence length       &  10   &  10      \\ 
Predicted future time steps &  1-40 &  1-40    \\ 
Hidden state of RNN         &  20   &  20      \\ 
Output dimension            &  2    &  1       \\ 
Number of RNN layer         &  1    &  1       \\ 
Dropout rate                &  0.2  &  0.2     \\ 
Epoch                       &  1000 &  1000    \\ 

\hline\hline
\end{tabular}

\label{tbl:DP_para}
\end{table}

\subsection{Performance Evaluation}
\label{subsec:pref}

\begin{figure}[t]
	\centering
	\includegraphics[width=.95\linewidth]{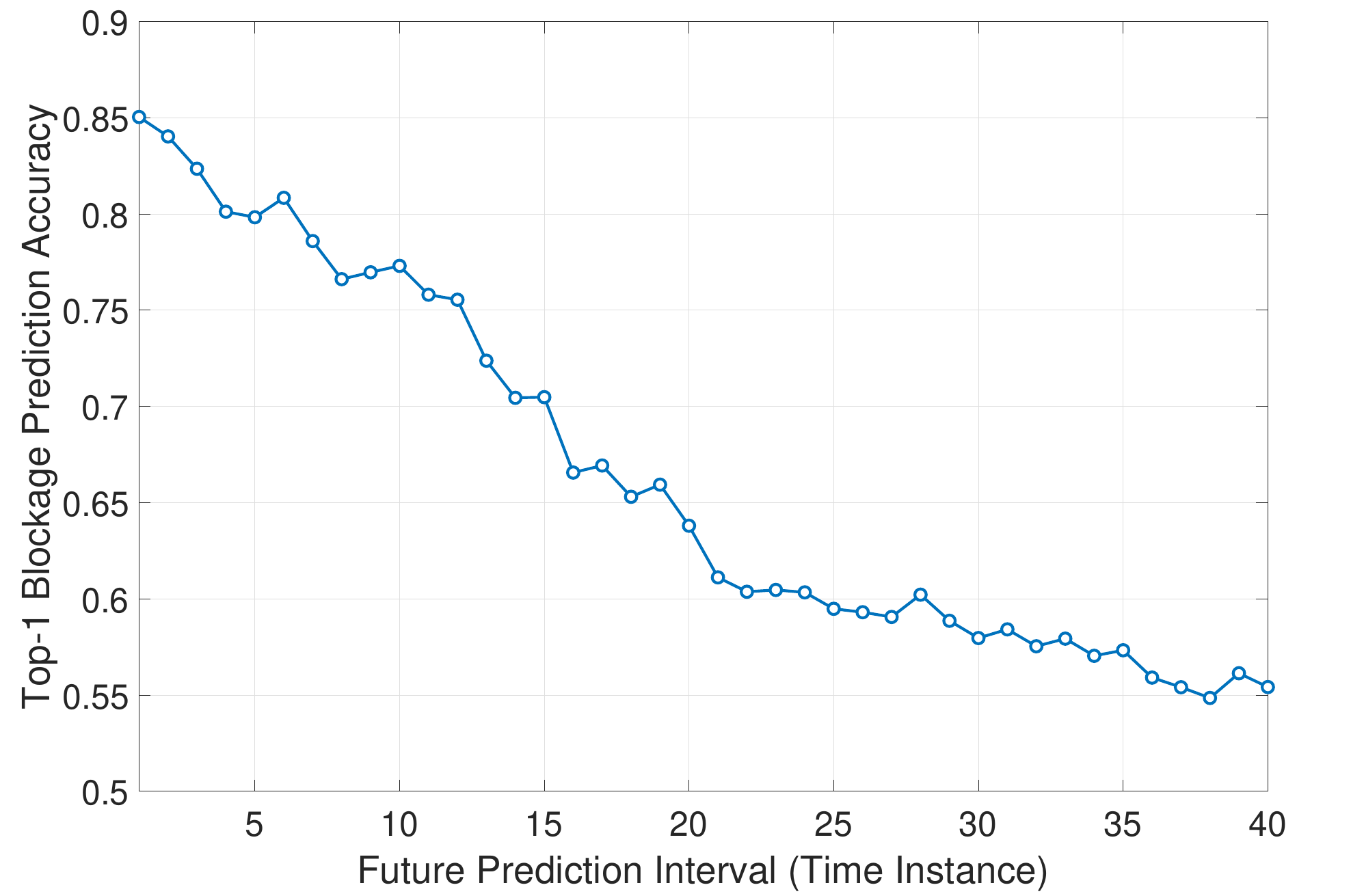}
	\caption{Top-1 blockage prediction accuracy for different future prediction intervals.}
	\label{fig:accuracy}
\end{figure}

\textbf{{Problem 1:}} Fig.~\ref{fig:accuracy} plots the Top-1 accuracy as a function of the prediction interval. We see that as the prediction interval increases the accuracy decreases; the Top-1 accuracy decreases sharply at first, and then flattens out. Our model achieves high accuracy when predicting the occurrence of blockage in near future. i.e. we can achieve above 80\% accuracy when predicting the occurrence of blockage in the future 6 time instances. This is because when the prediction interval is small, our training dataset contains a large number of sequences with a pre-blockage signature, resulting in a high ratio of these sequences in the training dataset. The prediction accuracy is good until the prediction interval goes beyond 15 time instances. It finally converges to the ``random guess'' performance as the prediction interval approaches 40 time instances, i.e., approaching an accuracy of $50\%$. Such general trend emphasizes value of pre-blockage signatures, which is commonly clear when the blockage is very close to the transmitter and receiver.

\begin{figure}
	\centering
	\includegraphics[width=.95\linewidth]{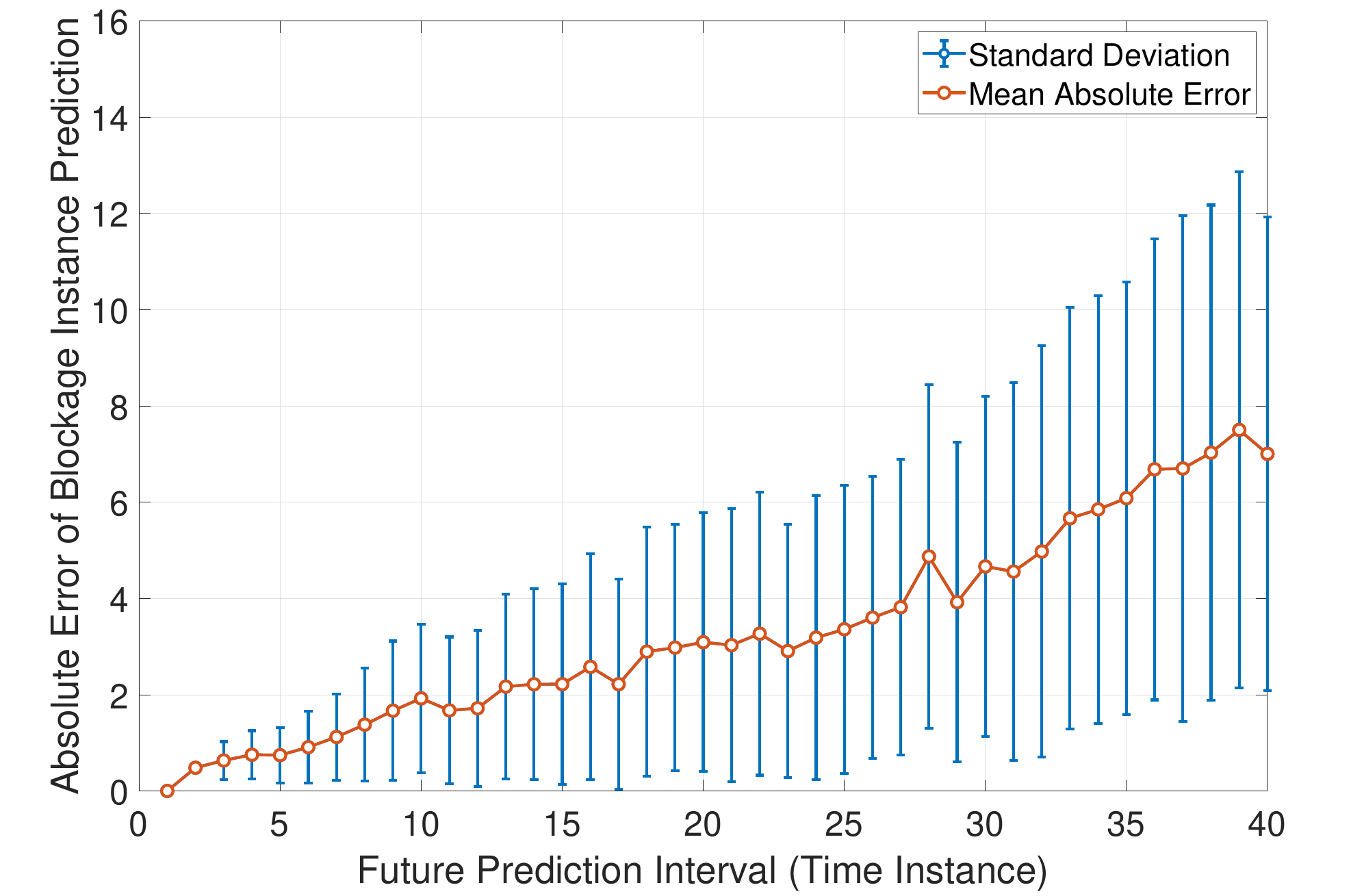}
	\caption{Mean absolute error between the target (the exact time instance where the blockage happens) and the prediction of this blockage time instance for different future prediction intervals.}
	\label{fig:avg_diff}
\end{figure}

\textbf{{Problem 2:}} Fig.~\ref{fig:avg_diff} plots the mean absolute error between prediction and ground-truth as a function of prediction interval. For each prediction instance, we show the standard deviation as an error bar. Given prediction interval below 10 time instance, our model can predict all the blockage transitions within 10 time instances under 1.9 with low volatility ($\pm$1.5). For a shorter prediction intervals, our observation window more likely falls into a pre-blockage signature resulting in lower prediction errors. However, as the prediction interval increases, our observation window captures more sequences without pre-blockage signature resulting in weaker prediction. However, when the prediction interval is 40 time instance, we can still predict the exact time of the blockage occurrence with the mean absolute error of under 8 time instances.

\section{Conclusion}
\label{sec:con}
In this paper, we proposed a deep-learning-based solution for the moving blockage prediction in mmWave communication systems. Specifically, we developed an RNN model to predict both the occurrence of the moving blockage and the exact time when the LOS link is blocked. Simulation results on real data showed that our model can achieve good performance for the moving blockage prediction, which allows the user to be proactively handed over to another base station without disconnecting the session with a high success probability.

\bibliographystyle{IEEEtran}
\bibliography{VinceRef}

\end{document}